\RequirePackage{letltxmacro}
\LetLtxMacro{\LaTeXtextbf}{\textbf}
\documentclass{ieeeaccess}
\LetLtxMacro{\textbf}{\LaTeXtextbf}

\IEEEoverridecommandlockouts

\usepackage{cite}
\usepackage{textcomp}

\usepackage{graphicx,subfigure}
\usepackage{amsmath,amssymb,amsfonts}
\usepackage{siunitx}
\usepackage{tabularx}
\usepackage{algorithmic}
\usepackage[linesnumbered,ruled,vlined]{algorithm2e}
\usepackage{amsthm}
\usepackage{graphicx,subfigure}
\usepackage{lscape}
\usepackage{verbatim}
\usepackage{color,soul}
\usepackage{xcolor}
\usepackage{subfigure}
\usepackage{tabularx,ragged2e,booktabs,caption,array,multirow,multicol}
\usepackage{csquotes}
\usepackage{mathtools}
\usepackage{lineno}
\usepackage{amsthm}  
\usepackage{listings}
\usepackage{latexsym}
\usepackage{epsfig}
\usepackage{float}
\usepackage{comment}
\usepackage{url}
\usepackage{xspace} 

\def\BibTeX{{\rm B\kern-.05em{\sc i\kern-.025em b}\kern-.08em
    T\kern-.1667em\lower.7ex\hbox{E}\kern-.125emX}}
\begin{document}
\history{Date of publication xxxx 00, 0000, date of current version xxxx 00, 0000.}
\doi{10.1109/ACCESS.2017.DOI}

\title{SMOTified-GAN for class imbalanced pattern classification problems}
\author{\uppercase{Anuraganand Sharma}\authorrefmark{1},
\uppercase{Prabhat Kumar Singh\authorrefmark{2}, and Rohitash Chandra}\authorrefmark{3}}
\address[1]{School of IT, Engineering, Mathematics \& Physics (STEMP), The University of the South Pacific, Fiji (e-mail: sharma\_au@usp.ac.fj)}
\address[2]{Department of Mechanical Engineering , IIT (BHU) Varanasi, India (e-mail: pksiit2000@gmail.com)}
\address[3]{School of Mathematics and Statistics, University of New South Wales, Sydney, Australia (email: rohitash.chandra@unsw.edu.au)}

\markboth
{Author \headeretal: Preparation of Papers for IEEE TRANSACTIONS and JOURNALS}
{Author \headeretal: Preparation of Papers for IEEE TRANSACTIONS and JOURNALS}

\corresp{Corresponding author: Anuraganand Sharma (e-mail: sharma\_au@usp.ac.fj).}

\begin{abstract}
Class imbalance in a dataset is a major problem for classifiers that results in poor prediction with a high true positive rate (TPR) but a low true negative rate (TNR) for a majority positive training dataset. Generally, the pre-processing technique of oversampling of minority class(es) are used to overcome this deficiency. Our focus is on using the hybridization of Generative Adversarial Network (GAN) and Synthetic Minority Over-Sampling Technique (SMOTE) to address class imbalanced problems. We propose a novel two-phase oversampling approach involving knowledge transfer that has the synergy of SMOTE and GAN. The unrealistic or overgeneralized samples of SMOTE are transformed into realistic distribution of data by GAN where there is not enough minority class data available for GAN to process them by itself effectively. We named it SMOTified-GAN as GAN works on pre-sampled minority data produced by SMOTE rather than randomly generating the samples itself. The experimental results prove the sample quality of minority class(es) has been improved in a variety of tested benchmark datasets. Its performance is improved by up to 9\% from the next best algorithm tested on F1-score measurements. Its time complexity is also reasonable which is around $O(N^2d^2T)$ for a sequential algorithm.
\end{abstract}


\titlepgskip=-15pt

\maketitle

\section{Introduction}
\label{S:1}


Class imbalance problem (CIP) refers to a type of classification problems where some classes are either majorly or moderately underrepresented in comparison to other classes \cite{ling_class_2010}. The unequal distribution makes many conventional machine learning algorithms quite less effective, especially for the prediction of minority classes \cite{castro_novel_2013}. A number of solutions have been proposed at the data and algorithm levels to deal with class imbalance such as preprocessing for oversampling or under-sampling, data augmentation, cost-sensitive learning/model penalization and one-class classification  \cite{ling_class_2010,makki_experimental_2019,haixiang_learning_2017,johnson_survey_2019}.

The imbalance dataset exhibits a major problem for the classifiers to be bias towards the majority class. The imbalanced class distribution results in the degradation of performance of the classifier model due to biased classification towards the majority class. It causes high true positive rate (TPR) and a low true negative rate (TNR) when majority samples are positive \cite{wei_role_2013}. Data imbalance can be commonly seen in fraud/fault/anomaly detection \cite{wei_effective_2013,lee_fault_2020, zhuo_gaussian_2020,huang_igan-ids_2020,makki_experimental_2019,mullick_generative_2019}, medical diagnosis of lethal and rare diseases \cite{johnson_survey_2019,bria_addressing_2020, qin_gan-based_2020}, software defect prediction \cite{software_defect}, natural disaster etc \cite{haixiang_learning_2017}. 

Commonly used pre-processing technique is oversampling as undersampling removes important information and does not result in accurate classification \cite{hilario_learning_2018}. Oversampling too suffers from inclusion of illegitimate samples which is still an active area of research \cite{zhu_improving_2020,tao_adaptive_2020}. Synthetic oversampling technique (SMOTE) \cite{chawla2002smote} is considered a ``de facto'' standard for an oversampling method. It is simple and effective; however, it may not produce diverse sample. SMOTE uses interpolation to randomly generate new samples from the nearest neighborhood of minority class data. It has been successfully used in regression \cite{torgo2013smote}, and classification problems \cite{jeatrakul2010classification} for a wide range of models \cite{li2020toxic}. A review of SMOTE and applications has been given in \cite{fernandez2018smote}.

The data samples in the case of imbalanced dataset can also be generated through classification models as well with data augmentation approach. Generative Adversarial Network (GAN) and its variations are commonly used to generate new ``fake'' samples \cite{zareapoor_oversampling_2021, DOUZAS2018464, ZHENG20201009, ZHAI2022313}. GAN was originally designed to generate the realistic-looking images for large datasets, however, it can also generate minority class samples thereby balancing the class distribution and avoiding over-fitting effectively \cite{zhang_imbalanced_2018}. Imbalanced data classification is ubiquitous in application domains. Data augmentation technique based on variations of GAN have been successfully applied on many applications such as skin lesion classification \cite{qin_gan-based_2020} for better diagnosis or pipeline leakage in petrochemical system \cite{xu_predicting_2019}.

Bayesian inference provides a principled framework to estimate unknown quantity represented by the posterior distribution  (parameters of a model) which is updated via Bayes' theorem  as more information gets available \cite{knill1996perception,box2011bayesian,andrieu2003introduction}. 
Markov Chain Monte Carlo (MCMC) sampling   is typically used to implement Bayesian inference \cite{neal2011mcmc}. It features   a likelihood function that  takes into account the prior distribution to either accept/reject samples obtained from a proposal distribution   to construct the posterior distribution of model parameters,    such as weights of a neural network     \cite{mackay1992practical,andrieu2003introduction,neal2011mcmc,van_ravenzwaaij_simple_2018}. 
 A major limitation for MCMC sampling technique is high computational complexity  for sampling from the    posterior   distribution \cite{das_racog_2015,johndrow_mcmc_2019}. There recently there has been much progress in MCMC sampling via the use of gradient-based proposals and parallel computing  in Bayesian deep learning \cite{chandra2019langevin,chandra2021revisiting,chandra2021bayesian}.  However, these have been mostly limited to model parameter (weights) uncertainty quantification rather than quantifying uncertainties in data or addressing class imbalanced problems. In the case of class imbalanced problems, MCMC sampling has been used for benchmark   real-world imbalanced datasets   \cite{das_racog_2015,das_wracog_2013}. MCMC method have been applied for handing imbalanced categorical data \cite{johndrow_mcmc_2019}.
Das et al. in \cite{das_racog_2015} have used Gibbs sampling (an MCMC method) to generate new minority class samples.

Another example of oversampling method is data dependant cost matrix, where a weighted misclassification cost is assigned to the misclassified classes \cite{haixiang_learning_2017}. It is not easy to determine this cost \cite{das_racog_2015}. The cost-sensitive loss function has penalty based weights for misclassification errors from both majority and minority classes. Hybrid neural network with a cost-sensitive support vector machine (hybrid NN-CSSVM) in \cite{kim_hybrid_2020} considers different cost related to each misclassification. Castro et al. in   \cite{castro_novel_2013} have improved the misclassification error for the imbalanced data by using the cost parameter according to the ratio of majority samples in the training set. One-class problem \cite{irigoien_towards_2014, lee_fault_2020, gao_handling_2020} also has a ``minority'' class but generally it is considered outlier which is removed from the training data. One-class modeling usually uses feature mapping or feature fitting to enforce the feature learning process \cite{gao_handling_2020}.

In this paper, we propose a novel hybrid approach that combines the strengths and overcomes the deficiency of two independent models that include SMOTE and GAN. SMOTE is known to produce some irregular or ``out of distribution'' samples. Additionally, SMOTE has not been generally used with deep learning and GAN generally has not been used for small datasets (minority classes) \cite{mullick_generative_2019}. Hence, we refer to it as SMOTified-GAN which is a two-phased process based on knowledge transfer or transfer learning \cite{pan_survey_2010}. Firstly, SMOTE generates promising samples which is then ``transferred'' to GAN which no longer uses random sample. Our approach may work well for both small and large datasets. This could lead  to more feasible and diverse data which are further enhanced through GAN to prepare better quality samples. We have obtained impressive results for our proposed method on numerical benchmark CIP datasets mainly from UCI library \cite{Dua:2019}. Its efficiency is also reasonable which is the combination of SMOTE and GAN as discussed in Section 2 and Section 3. SMOTified-GAN, however, works on non-image data only in its current form.

The rest of the paper is organised as follows. Section 2 presents the state-of-the-art techniques to solve CIPs. Section 3 discusses the proposed method -- SMOTified-GAN. Section 4 shows the experimental results and Section 5 discusses the outcome of the experiments. Lastly, Section 6 concludes the paper by summarizing the results and proposing some further extensions to the research.   

\section{Related Work on class imbalance problems}

\subsection{Synthetic Minority Oversampling TEchnique (SMOTE)}

The SMOTE is a ``de facto'' standard for pre-processing imbalanced data. This is not a complete random sampling whereas it uses interpolation among the neighboring minority class examples. It is efficient and easy to implement. Each minority example gets $k$-nearest neighbors (KNN) which are randomly selected to have interpolation to create new samples. The pseudocode is given in \algorithmcfname{} \ref{Algo:SMOTE}. The parameters $n$ and $d$ are the size and dimension of the minority class respectively; $N$ is the size of the majority class and parameter $k$ for $k$-nearest neighbor. Lines 1-5 finds KNN for each minority sample then does the interpolation with them to create new samples. Lines 6-12 describes the interpolation step where $N-n$ samples are being created and added into minority class. Its time complexity for a single machine has the order of $O((N-n)dn{\log k}) \approx O(N^2d{\log k})$ \cite{raschka_stat_2018, fernandez_smote_2018, chawla_smote_2002}.

\begin{algorithm}
    \SetAlgoLined
    \caption{Pseudocode for SMOTE}
    \label{Algo:SMOTE}
    \tcp{Input: $d$-dimensional minority samples $X$ of size $n$ from a training data set of size $N$ that requires $N-n$ over-samples. $k$ defines $k$-nearest neighbors.}
    $N \leftarrow N-n$\\
    \For{$i = 1:\Arrowvert X \Arrowvert$}{
    $S \leftarrow$ KNN($x_i,k$) \tcp{$x_i \in X$}
    $X \leftarrow$ interpolate($N/100, x_i, S$) \tcp{for $N>100$}
    }
    \BlankLine
    \tcp{subroutine for interpolate}
    \SetKwFunction{Fpopulate}{interpolate}
    \Fpopulate ($N, x_i, S$){\\
    \While{$\Arrowvert X \Arrowvert<N$}{
    $a \leftarrow R_I^{k \times 1} (1)$ \tcp{pick a random integer value from 1...k}
    $x_j \leftarrow S\{a\}$\\
    \tcp{$\Delta x_{ij} = x_j-x_i \Rightarrow$ euclidean distance between $x_i$ and $x_j$}
    \tcp{$R_D^{1 \times d} \Rightarrow$ a decimal random number between 0 to 1}
    $X \leftarrow X \bigcup (x_i + \Delta x_{ij} \times R_D^{1 \times d})$
    }
    \KwRet X
    }
    
\end{algorithm}

There are many variants of SMOTE that have been successfully applied to various application domains such as bioinformatics, video surveillance, fault detection or high dimensional gene expression data sets \cite{won_push_2020,fernandez_smote_2018, blagus_evaluation_2012}. There are many variants of SMOTE such as regular SMOTE, Borderline-SMOTE, SVM-SMOTE and KMeans-SMOTE \cite{zheng_smote_2020}. Kovacs in \cite{kovacs_smote-variants_2019} has shown the implementation of 85 variants of SMOTE in python library.

\subsection{Generative Adversarial Network (GAN)}
 GAN is a class of machine learning frameworks in which there is a contest between two neural networks with a continuous and simultaneous improvement of both neural networks. This technique learns to generate new data with the same statistics as the training set by capturing the true data distribution \cite{goodfellow_generative_2014}.

GAN has been successfully used for data augmentation. The two neural networks of GAN learn the target distribution and generate new samples to achieve similar distributive structure in its generated over-sampled data. A GAN is simply the synergy of two deep learning network that produce "fake" data examples emulating the properties of the real data \cite{zhang_imbalanced_2018,suh_cegan_2021,ali-gombe_mfc-gan_2019}.

GAN had not been designed for oversampling imbalanced classes but to create ``fake'' images of real images which should be hard to distinguish. However, its success in data augmentation for over-sampling has led to the introduction to many variations of GAN to solve CIP \cite{goodfellow_generative_2014,sorin_creating_2020,lin_fpgan_2021,yi_progressive_2020, DOUZAS2018464, ZHENG20201009, ZHAI2022313}.

\begin{figure}[th]
	\begin{center}
		\includegraphics[scale=0.40]{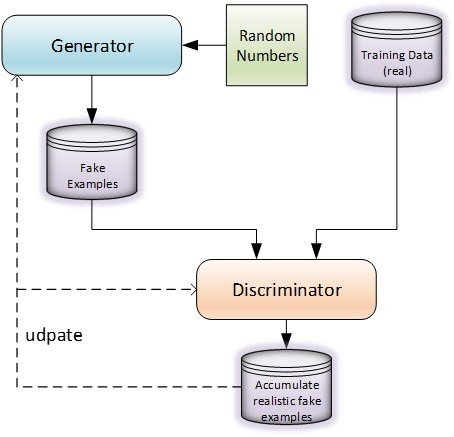}
	\end{center}
	\vspace{-5mm}
	\caption{Process of ``fake'' sample generation with GAN}
	\label{fig:gan}
\end{figure}
 
The first network is called Generator whose responsibility is to takes a vector of random values and generate the data similar to the real data used in training. The second network is called Discriminator that takes input data from both the real training data and the ``fake'' data from the generator, to classify them correctly. This process is shown in \figurename{} \ref{fig:gan}.

The time-complexity of GAN can be roughly given as $O(nTLd^2)$ where the new parameters $L$ and $T$ are layer-size and total iterations for a GAN. Its convergence rate with the Stochastic Gradient Descent would be $O(\frac{1}{T} + \sigma^2)$ where $\sigma^2$ is the variance of the dataset \cite{sharma_guided_2021}. See Section \ref{Section:SMOTified-GAN} for further details.

\section{SMOTified-GAN for Class Imbalance Problem} 
\label{Section:SMOTified-GAN} 
\begin{figure*}[th]
	\begin{center}
		\includegraphics[scale=0.50]{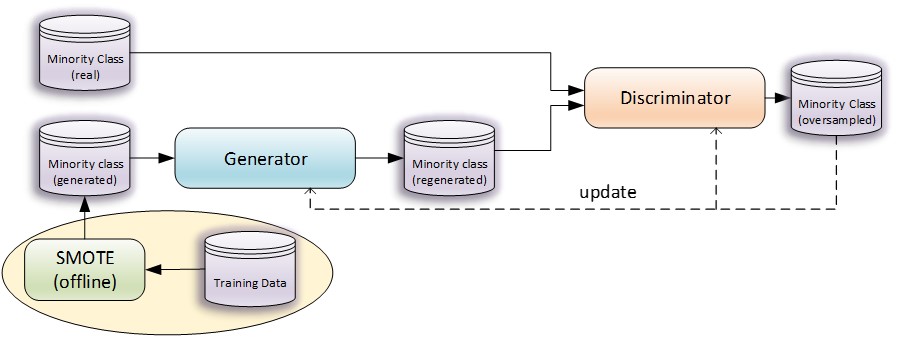}
	\end{center}
	\caption{Process of sample generation with SMOTified-GAN}  
	\label{fig:smotified_gan}
\end{figure*}

Our proposed method tries to overcome the deficiency of both SMOTE and GAN by using a transfer learning concept where it first extracts the knowledge about minority class from SMOTE and then applies it to GAN. We have named it SMOTified-GAN as it tries to diversify the original samples produced by SMOTE through GAN. Additionally, the quality of the sample is further enhanced by emulating them with the realistic samples. The process of SMOTified-GAN is shown in \figurename{} \ref{fig:smotified_gan}.  

\begin{figure}[th]
    \centering
    \subfigure[Low-diversity with SMOTE taken from  \cite{jefferson_bank_2020}]{
        \includegraphics[scale=0.15]{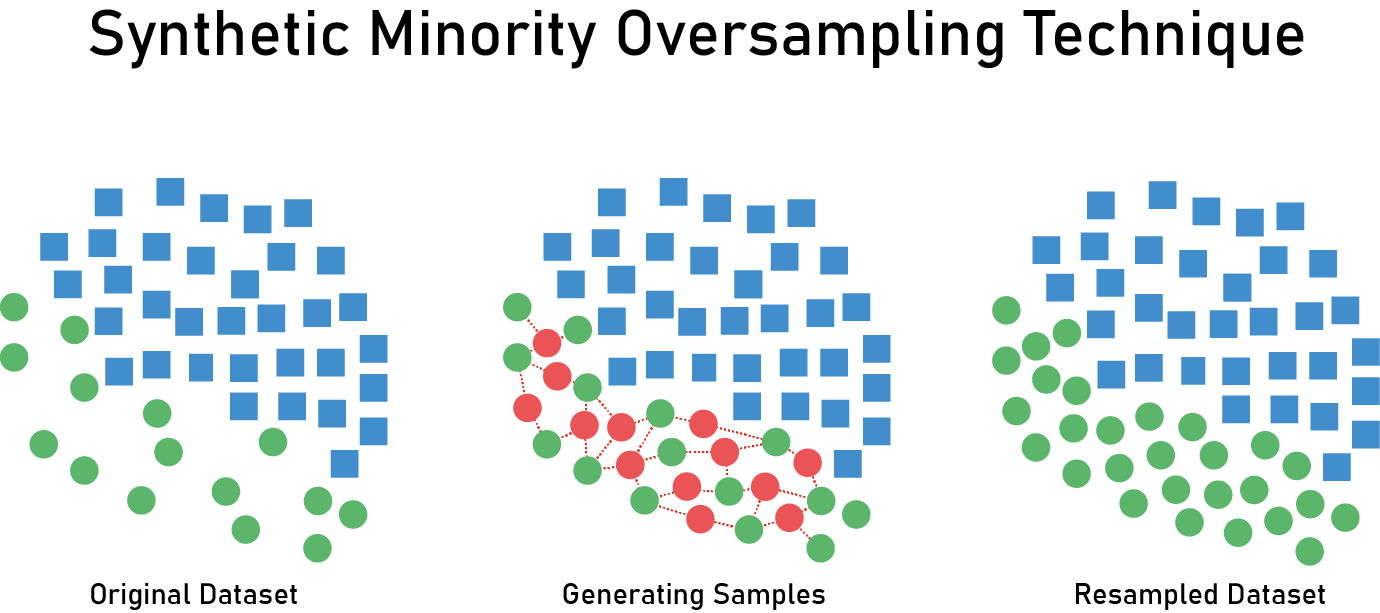}
        \label{fig:smote_drawback_a}
    }
    \hfill
     \subfigure[Interpolation with SMOTE taken from \cite{hu_novel_2013}]{
        \includegraphics[scale=1.20]{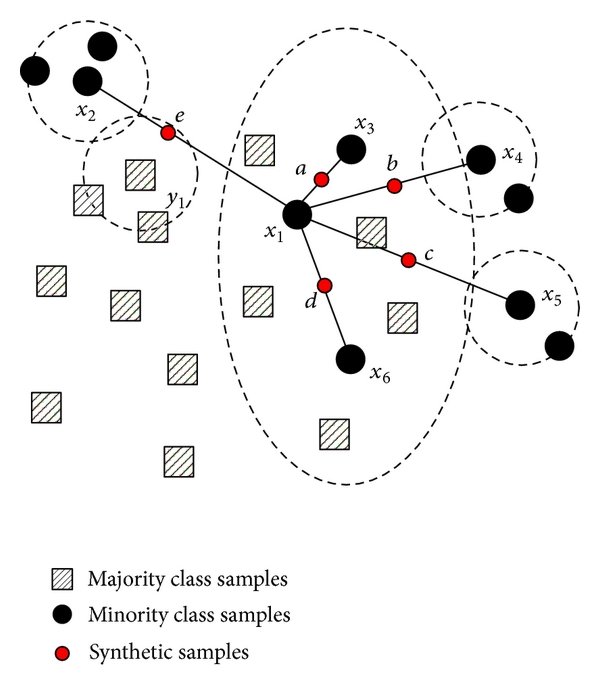}
        \label{fig:smote_drawback_b}
    }
    \hfill
    \caption{SMOTE processing for oversampling}
    \label{fig:smote_drawback}
\end{figure}

Even though SMOTE is widely used as an oversampling technique, it suffers with some deficiency. The major drawback of SMOTE is that it focuses on local information and therefore it does not generate diverse set of data as shown in \figurename{} \ref{fig:smote_drawback_a}. Additionally, \figurename{} \ref{fig:smote_drawback_b} shows the 5 nearest neighbors of $x_1$, $\{x_2,...,x_6\}$ are firstly, blindly chosen then interpolated (using Euclidean distance) to get the corresponding synthetic samples $\{a,...,e\}$. Even, there there is a high chance of miss-classification for sample $e$ with a majority sample $y_1$ \cite{hu_novel_2013}. The generated data are generally insufficiently realistic compared to GAN that captures the true data distribution in order to generate data for the minority class \cite{zheng_conditional_2020}.

GAN  is not ideally fit for oversampling as it has been originally designed for realistic looking images with convolutional neural networks (CNN) rather than producing over-samples for the minority class. Additionally, GAN may face data scarcity problem as minority class is already in reduced form where model training requires more of its data to be sacrificed for validation and testing purpose. Though, cross-validation techniques may solve this problem to some extent. 

The architecture of GAN consists of two networks as mentioned in the previous section where the objective of the generator network is to generate data that fools the discriminator network to classifies as ``real''.  To optimize its performance, maximize the loss of the discriminator when data is coming from the generator. The objective of the generator is to generate data that the discriminator classifies as ``real''. To optimize the performance of the discriminator, the loss of the discriminator is to be minimized when given batches of both real and generated data. The objective of the discriminator is to not be ``fooled'' by the generator \cite{noauthor_train_2021,goodfellow_generative_2014}.

The discriminator score can be given as:

\[\max_D\mathbb{E}_x [\log D(x)]+\mathbb{E}_z[\log  (1-D(G(z)))] \]

or 
\[\min_D\mathbb{E}_x [-\log D(x)]-\mathbb{E}_z[\log  (1-D(G(z)))] \]

 $D(x)$ contains the discriminator output probabilities for the real data $x$ and $D(G(z))$ contains the discriminator output probabilities for the generated data $z$.
 
The generator score is:
 \[\min_G-\mathbb{E}_z[\log D(G(z))] \]

The pseudocode for the GAN algorithm is given in \algorithmcfname{} \ref{Algo:GAN} where $SGD$ and $weights$ are functions to determine gradient for a mini-batch using Stochastic Gradient Descent algorithm (SGD) optimizer \cite{sharma_guided_2018} or its any other variation such as ADAM \cite{kingma_adam:_2014} or RMSprop \cite{zeiler_adadelta:_2012}, and update the weights respectively. Once the algorithm terminates `good' fake samples are collected with \textit{accumulateFakeEx} based on classification accuracy. 

\begin{algorithm}
    \SetAlgoLined
    \caption{Pseudocode for GAN}
    \label{Algo:GAN}
     \tcp{Input: training data set examples $x$ and noise samples $z$ from appropriate random number generator. An optional parameter can be the size $n_{fake}$ of fake sample needed.}
    \tcp{initialize parameters \\ $m_i$ is minibatch indices for $i^{th}$ index and $T$ is total iterations.}
    \SetKwFunction{FGAN}{GAN}
    \FGAN ($x, z, n_{fake}$){\\
    \For{$t = 1:T$}{
    \tcp{generally step size S is 1}
    \tcp{subscript $d$ and $g$ refers to discriminator and generator entity respectively}
    \For{$s = 1:S$}{ 
    $g_d \leftarrow SGD(-\log D(x)-\log  (1-D(G(z)), W_d, m_i)$\\ 
    $W_d \leftarrow weights(g_d, W_d)$\\
    }
    $g_g \leftarrow SGD(-\log D(G(z)), W_g, m_i)$\\
    $W_g \leftarrow weights(g_g, W_g)$\\
    }
    $x' \leftarrow $ accumulateFakeEx ($Model_d(W_d,x,z)$,      $Model_g(W_g,x,z), n_{fake}$)\\
    \KwRet $x'$
    }
\end{algorithm}

 Goodfellow   \cite{NIPS2014_IanG} has used \textit{sigmoid}   as the activation function that would result the following scores to minimize:

Discriminator:
\[\mathbb{E}_x [-\log  (1+e^{-y})]  - \mathbb{E}_z [1-\log  (1+e^{-\hat{y}})] \]
 
Generator:
\[\mathbb{E}_z[-\log  (1+e^{-\hat{y}})] \]

where $y$ and $\hat{y}$ are the outputs of the  Discriminator D and Generator model G respectively before the activation function is applied.

The formalization of SMOTified-GAN is not very different from the original GAN. Only the random generator function of GAN is replaced with the repertoire of oversample minority examples from SMOTE. The modified scores can be shown as:

discriminator score: 
\[\max_D\mathbb{E}_{x^*} [\log D(x^*|x)]+\mathbb{E}_u[\log  (1-D(G(u)))] \] 

Generator score:
\[\min_G-\mathbb{E}_u[\log D(G(u))] \]

where $x^*$ is training samples of minority class(es) and $u$ is over-sampled data of the same class(es) generated from different algorithms such as SMOTE in this case. The pseudocode for SMOTified-GAN is given in \algorithmcfname{} \ref{Algo:SMOTified-GAN}. Its implementation is not too difficult either. The Python code is available at \url{ https://github.com/anuraganands}.

As illustrated in \figurename{} \ref{fig:smotified_gan}, there are two sections of SMOTified-GAN. The first one replaces the random number generator (refer \figurename{} \ref{fig:gan}) with the repertoire of oversamples from SMOTE. The second section continues with the process of GAN using the new samples from SMOTE. \algorithmcfname{} \ref{Algo:SMOTified-GAN} also shows this process in two steps. Line  (1) calls SMOTE function given in \algorithmcfname{} \ref{Algo:SMOTE} and then Line (2) calls GAN function given in \algorithmcfname{} \ref{Algo:GAN}. However, this time the generated samples $u$ is used instead of random noise $z$.

\begin{algorithm}
    \SetAlgoLined
    \caption{Pseudocode for SMOTified-GAN}
    \label{Algo:SMOTified-GAN}
     \tcp{Input: minority examples $x*$ from a training data set $x$ of size $N$ that requires $N-n$ over-samples;\\
     User-defined parameter $k$ for $k$-nearest neighbors.}
    \BlankLine
    \tcp{First execute SMOTE given in \algorithmcfname{} \ref{Algo:SMOTE} then GAN given in \algorithmcfname{} \ref{Algo:GAN}}
    $u \leftarrow $ call \algorithmcfname{}\_\ref{Algo:SMOTE} ($x*,k$) \tcp{generate over-sampled minority examples $u$.}
    $u \leftarrow $ call \algorithmcfname{}\_\ref{Algo:GAN} ($x*, u, N-n$). 

\end{algorithm}

Its time complexity for sequential algorithm is combination of SMOTE's and GAN's time complexity, i.e., $O(N^2d{\log k} + nTLd^2) \approx O(N^2d+TNd^2)$ Since $n$ is a small part of $N$ so it can be assumed $nL$ is comparable to $N$. This can further simplify the complexity to $O(N^2d+TNd^2) \leq O(N^2d^2 (1/d+T/N)) \leq O(N^2d^2T)$.

The major difference between our proposed method SMOTEfied-GAN and GAN is the use of ready-made repertoire of samples generated from SMOTE instead of a set of random noise to begin with. Intuitively, this helps in improvement of the input samples that produces better over-samples. This natural synergy of SMOTE and GAN guides the na\"ive GAN to have a jump-start with promising data before going through further refinement of unrealistic data from SMOTE.


 \section{Experiments and Results}
 In this section, we provide experimental results of over-sampling methods, namely, SMOTE, GAN and SMOTified-GAN on different datasets that have been taken from the literature of CIP \cite{bhowan_reusing_2014,nunez_improving_2017, napierala_types_2016}. The over-sampled minority class data have been made equal to the majority class data which is then augmented into training data that are then fed into the Neural Networks (NN) for classification. We have also done the testing on original datasets without using any data augmentation technique. 
 
\subsection{Datasets}

 We evaluate and compare our model on small to large datasets that feature class imbalance as shown in \tablename{} \ref{tab:dataset}. The datasets were mainly obtained from the UCI machine learning repository \cite{Dua:2019} \vphantom{and KEEL-dataset repository cite{alcala2011keel} and}that have been used in a number of methods for CIPs \cite{bhowan_reusing_2014,nunez_improving_2017, napierala_types_2016}. Some datasets such as Credit-Card Fraud and Shuttle are highly imbalanced with minority class contribution as 0.17\% and 0.29\% respectively. 

\begin{table*}[htbp!]
\small
\caption{Description for small to large datasets}
\centering
 \begin{tabular}{l  l   l   l   l   l} 
 \hline
  \hline
 Dataset &  Features   &  Classes &  Instances & Minority Class (\%) & Description\\
 \hline
 \hline
 Ecoli \cite{Dua:2019} & 7 & 2 & 335 & 5.97 &  This data contains protein localization sites  \\ 
 \hline
 Ionosphere \cite{Dua:2019} & 34 & 2 & 351 & 35.71 &Classification of radar returns from the ionosphere
 \\ 
 \hline
 Pageblocks \cite{Dua:2019} & 10 & 2 & 471 & 5.94 & Classifying all the blocks of the page layout of a document \\ 
   \hline
 Yeast \cite{Dua:2019} & 8 & 2 & 513 & 9.94 &  Predicting the Cellular Localization Sites of Proteins \\ 
   \hline
 Wine \cite{Dua:2019} & 11 & 2 & 655 & 2.74 & Using chemical analysis determine the origin of wines  \\ 
   \hline
 Poker \cite{Dua:2019} & 10 & 2 & 1476 & 1.15 & Purpose is to predict poker hands  \\ 
   \hline
 Abalone \cite{Dua:2019} & 8 & 2 & 4177 & 20.1 & Predict the age of abalone from physical measurements \\ 
    \hline  
 Spambase \cite{Dua:2019} & 57 & 2 & 4601 & 39.39 & Classifying Email as Spam or Non-Spam  \\ 
    \hline
 Shuttle \cite{Dua:2019} & 9 & 2 & 58000 & 0.294 &  Approximately 80 percent of the data belongs to class 1 \\ 
 \hline
 Credit-Card Fraud \cite{noauthor_credit_nodate}& 30 & 2 & 284807 & 0.172 &
This dataset has 492 frauds out of 284,807 transactions. \\ 
  \hline
 Connect4 \cite{Dua:2019} & 42 & 2 & 376640 & 3.84 &  Contains connect-4 positions  \\ 
  \hline  
   \hline
\end{tabular}
\label{tab:dataset}
\end{table*}
 
\subsection{Experimental Setup}
\sethlcolor{pink}

\begin{figure}[th]
  \begin{center}  
   \includegraphics[scale=0.80]{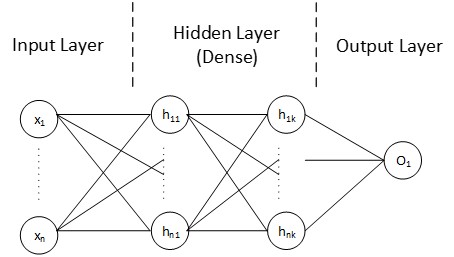}
    \caption{Architecture of the classifier}
 \label{fig:architecture}
  \end{center}
\end{figure}

We used na\"ive GAN model \cite{goodfellow_generative_2014} and na\"ive SMOTE \cite{chawla_smote_2002} in this paper. SMOTified-GAN  uses the above two models, however, it is flexible enough to work with other combinations of different variations as well. The parameter settings such as learning rate, total epochs and loss functions are shown in \tablename{} \ref{tab:paraSet}. The GAN generator neural network features 3 hidden layers with 128 neurons in each layer. The GAN discriminator network is similar to the generator network with major difference of having only two layers first a linear layer followed by a leaky-ReLu layer with alpha=0.2. The classifier architecture is given in \figurename{} \ref{fig:architecture}. In GAN training, we use binary cross-entropy activation function with training data batch-size of 128 and initial learning-rate of 0.00001 with Adam optimser. 

After basic pre-processing steps, SMOTE oversampling is done with $k=5$ neighbors. The stopping criteria for SMOTified-GAN and na\"ive GAN's training are based on validation error to avoid any over-learn. Additionally, it is ensured that the discriminator and generator loss remain significant and do not approach near zero.
 


\begin{table*}[htbp!]
\small
\caption{Parameter Settings }
\centering
 \begin{tabular}{l  l   l   l}
 \hline
  \hline
 Parameter & Neural Network & Generator & Discriminator \\
 \hline
   \hline
Total neurons per hidden layer: & 256, 128 & 128, 256, 512, 1024 & 512, 256, 128\\  
\hline  
  
Optimizer : & Adam & Adam & Adam\\  
  \hline
Loss Function : & Mean Absolute Error & BCEWithLogitsLoss & BCEWithLogitsLoss\\ 
  \hline
Activation : & ReLU & ReLU & LeakyReLU (0.2)\\ 
  \hline
Normalization : & - & BatchNorm1d & -\\ 
  \hline
Learning Rate : & 0.00001 & 0.00001 & 0.00001\\ 
  \hline
   \hline
\end{tabular}
\label{tab:paraSet}
\end{table*}

\subsection{Preliminary investigation}


The experiment has been conducted on 11 benchmark imbalanced datasets that are trained on NN to test the efficacy of various oversampling techniques. We used SMOTE, GAN and our proposed method SMOTified-GAN for oversampling. We have also done the testing with original data without any data augmentation. The quality of classification and comparative results are shown in 
\tablename{} \ref{tab:results}. As expected all datasets show high train and test accuracy due to high imbalance in the datasets. So it is important to look into F1 scores to determine high precision and recall measures. The best F1 scores have been shown with the bold font. 

It is clear from the experimental results that SMOTified-GAN has outperformed other oversampling techniques. Only Connect4 is an outlier where all oversampling techniques are showing poor results compared to the non-oversampling technique. Surprisingly, SMOTE also performed poorly by 3.6\% compared to the original training dataset without any data augmentation. This result can be attributed to the fact that the dataset is highly imbalanced where minority class constitutes only 3.84\% of the training dataset. This does not provide enough data for generalization. So the minority class should not be over-sampled blindly for a given dataset. Conversely, no data augmentation with datasets such Ecoli (6.0\% minority class) and Wine (2.7\% minority class) shows very poor and unacceptable results. Here data augmentation techniques especially with SMOTified-GAN show much better results of F1-score of 92.2\% and 52.7\% respectively.

\begin{figure}[th]
  \begin{center}  
   \includegraphics[scale=0.9]{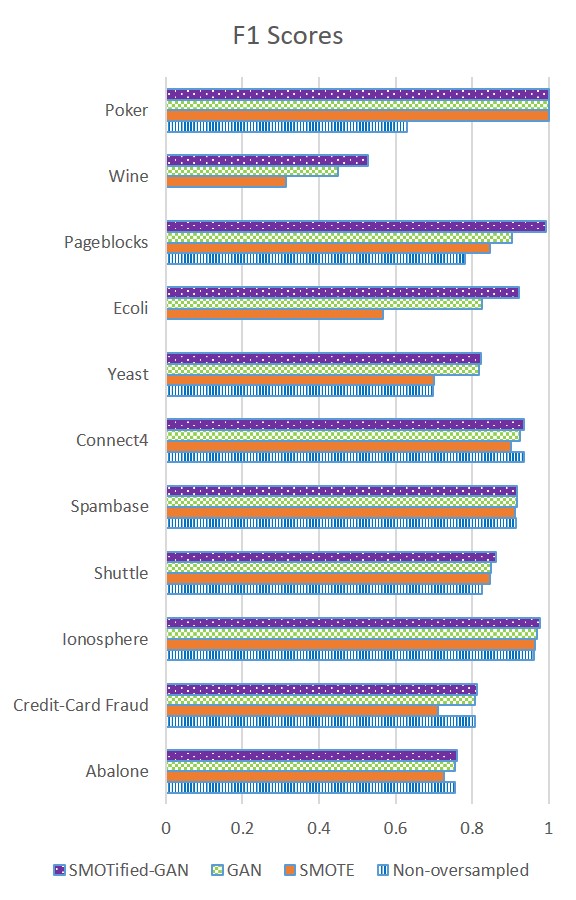}
    \caption{Comparison of F1 scores}
 \label{fig:F1score}
  \end{center}
\end{figure}

SMOTified-GAN gives the best results -- considering F1 score -- for all other datasets with the diverse proportion of minority class such as Creditcard Fraud  (0.2\% minority class), Spambase (39.4\% minority class), Yeast (9.9\% minority class) and Wine (2.7\% minority class). The rest of the datasets, Ionosphere, Shuttle, Ecoli, Pageblocks and Poker also favors SMOTified-GAN. \figurename{} \ref{fig:F1score} on the comparative F1 score shows SMOTified-GAN outperforms other algorithms on 10/11 datasets. Its performance is significantly improved for Pageblocks by 9\% and 10\% for Ecoli. SMOTified-GAN has also produced better precision and recall for most of the datasets. It has the best precision for all 11 datasets and the best recall for 9/11 datasets. GAN and SMOTE give mixed results on different datasets. GAN has produced 10/11 times better results than SMOTE. Notably, data augmentation less training is also better than GAN and SMOTE with 2/11 times and 4/11 times respectively. 

Datasets like Yeast, Ecoli, Wine, Poker and Pageblocks have small number of minority class data instances relative to the majority class which allows SMOTified-GAN to show its  potential over other algorithms as seen in the respective results. In datasets like Ecoli and Wine the minority instances are so low that the non-oversampling method completely fails to predict the minority class. All models give high train and test accuracy on all datasets which is attributed to the dominance of majority class in these datasets hence the true performance index measure is the minority F1-score which depends on both the precision and recall. Overall, the proposed model of SMOTified-GAN outperforms the others in terms of F1-score and a comparatively low standard deviation of results.

\begin{figure*}[htbp!]\centering
\centering     
\subfigure[abalone-1]{\label{fig:SMOTified-GAN_loss:a}\includegraphics[width=50mm]{images/images of  curves/aballoss1.png}}
\subfigure[creditcard]{\label{fig:SMOTified-GAN_loss:b}\includegraphics[width=50mm]{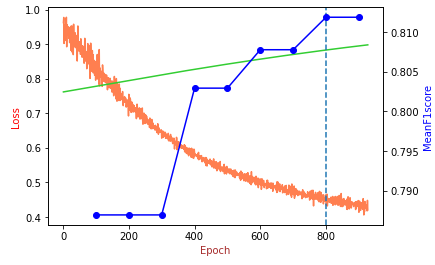}}
\subfigure[ecoli]{\label{fig:SMOTified-GAN_loss:c}\includegraphics[width=50mm]{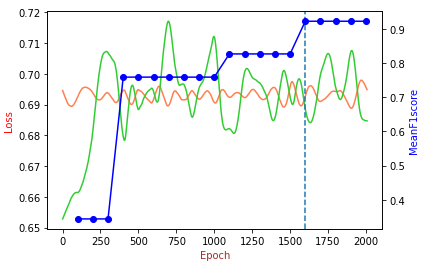}}
\subfigure[ionosphere]{\label{fig:SMOTified-GAN_loss:d}\includegraphics[width=50mm]{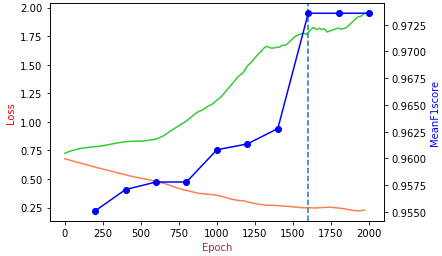}}
\subfigure[pageblocks]{\label{fig:SMOTified-GAN_loss:e}\includegraphics[width=50mm]{images/images of  curves/pageloss1.png}}
\subfigure[poker]{\label{fig:SMOTified-GAN_loss:f}\includegraphics[width=50mm]{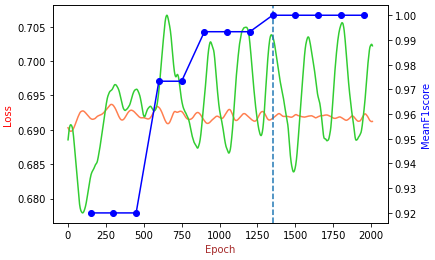}}
\subfigure{\label{fig:SMOTified-GAN_loss:legends}\includegraphics[width=80mm]{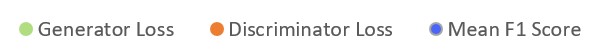}}

\caption{SMOTified-GAN's Loss and epoch for the best F1-score}
\label{fig:SMOTified-GAN_loss}
\end{figure*}

The training loss curves of SMOTified-GAN's generator and discriminator models w.r.t the number of epochs during training of selected datasets have been shown in \figurename{}  \ref{fig:SMOTified-GAN_loss}. In general, the Discriminator's loss curve converges fairly quickly whereas the Generative loss curve demonstrates high fluctuations, however, these fluctuations generally gets steady at around 2000 epochs. We have also drawn validation F1-score in the same graph to determine the termination criterion. The training stops once the validation F1-score reaches its highest value to avoid any over-fitting.

\subsection{Results}
Table \ref{tab:results} presents a summary for the experimental results with NN using the respective oversampling methods -- SMOTE, GAN, SMOTified-GAN and also without no augmentation. It shows training and test accuracy and measurements of F1-score, precision and recall. Its purpose is to demonstrate the effectiveness of the methods for class imbalanced datasets. We report the mean, standard deviation, and best performance using the respective evaluation metrics using 30 experimental runs where each run has a different randomised initial position in weight space. This is done to incorporate model uncertainty in our results.

\begin{table*}[htbp!]
\small
\caption{Comparison of experimental results on NN with baseline methods (SMOTE, GAN, SMOTified-GAN and non-oversampled original data) }
\centering
 \begin{tabular}{p{1.7cm}  p{2.3cm}   c   c   c   c   c } 
 \hline
 \hline
 Dataset & Oversampling & Train  & Test  & F1  & Precision   & Recall\\
   & methods & Mean (Best, SD)  & Mean (Best, SD)  & Mean (Best, SD)  &    & \\
   \hline  
   \hline
 Ecoli & Non-oversampled & 0.9328 (0.9328,0.0000)& 0.9701 (0.9701,0.0000) & 0.0000 (0.0000,0.0000) & 0.00 & 0.00 \\ 
  
  & SMOTE & 0.9905 (0.9959,0.0020) & 0.9577 (0.9701,0.0100) & 0.5684 (0.6666,0.0890) &0.50 & 1.00 \\

  & GAN & 0.9885 (0.9919,0.0013) & 0.9880 (1.0000,0.0099) & 0.8266 (1.0000,0.1964) &1.00 & 1.00 \\  

 & SMOTified-GAN & 0.9861 (0.9879,0.0010) & 0.9960 (1.0000,0.0077) & \textbf{0.9222 (1.0000,0.1433)} & 1.00 & 1.00 \\
   \hline
 Ionosphere & Non-oversampled & 0.9878 (0.9928,0.0027)& 0.9728 (0.9857,0.0126) & 0.9621 (0.9803,0.0179) & 0.98 & 0.98 \\ 
  
  & SMOTE & 0.9914 (0.9916,0.0007) & 0.9738 (0.9857,0.0113) & 0.9632 (0.9803,0.0164) & 0.97 & 0.99 \\

  & GAN & 0.9901 (0.9944,0.0017) & 0.9767 (1.0000,0.0086) & 0.9701 (1.0000,0.0210) &1.00 & 1.00 \\  
  
 & SMOTified-GAN & 0.9903 (0.9944,0.0023) & 0.9823 (1.0000,0.0068) & \textbf{0.9777 (1.0000,0.0169)} & 1.00 & 1.00 \\
 
   \hline  
 Pageblocks & Non-oversampled & 0.9627 (0.9627,0.0000)& 0.9775 (0.9894,0.0050) & 0.7803 (0.9090,0.0700) & 1.00 & 0.82 \\ 
  
  & SMOTE & 0.9955 (1.0000,0.0030) & 0.9761 (1.0000,0.0130) & 0.8480 (1.0000,0.0780) &1.00 & 1.00 \\

  & GAN & 0.9943 (0.9972,0.0020) & 0.9858 (1.0000,0.0098) & 0.9038 (1.0000,0.0793) &1.00 & 1.00 \\ 
 
 & SMOTified-GAN & 0.9943 (1.0000,0.0043) & 0.9989 (1.0000,0.0042) & \textbf{0.9926 (1.0000,0.0291)} & 1.00 & 1.00 \\
 
 \hline  
 Yeast & Non-oversampled & 0.9748 (0.9780,0.0013)& 0.9323 (0.9417,0.0097) & 0.6987 (0.7272,0.0305) & 0.80 & 0.67\\ 
 
  & SMOTE & 0.9638 (0.9757,0.0083) & 0.9139 (0.9417,0.0152) & 0.7012 (0.7857,0.0446) & 0.82 & 0.75 \\
  
  & GAN & 0.9782 (0.9811,0.0028) & 0.9595 (0.9514,0.0036) & 0.8173 (0.8333,0.0169) & 0.83 & 0.83 \\  
 
 & SMOTified-GAN & 0.9663 (0.9703,0.0023) & 0.9611 (0.9611,0.0044) & \textbf{0.8221 (0.8695,0.0223)} & \textbf{0.91} & \textbf{0.83} \\
 
  \hline    
 Wine & Non-oversampled & 0.9770 (0.9770,0.0000)& 0.9541 (0.9541,0.0000) & 0.0000 (0.0000,0.0000) & 0.00 & 0.00 \\ 
  
  & SMOTE & 0.9806 (0.9843,0.0020) & 0.9081 (0.9389,0.0090) & 0.3149 (0.5000,0.0651) &0.39 & 0.67 \\

  & GAN & 0.9841 (0.9873,0.0020) & 0.9549 (0.9618,0.0073) & 0.4489 (0.5454,0.1112) &0.46 & 0.67 \\  
 
 & SMOTified-GAN & 0.9854 (0.9873,0.0010) & 0.9558 (0.9694,0.0090) & \textbf{0.5274 (0.6000,0.0780)} & \textbf{0.53} & \textbf{0.69} \\
 
 \hline 
 Poker & Non-oversampled & 0.9902 (0.9906,0.0008)& 0.9949 (0.9966,0.0020) & 0.6300 (0.8000,0.2530) & 1.00 & 0.67 \\ 
  
  & SMOTE & 1.0000 (1.0000,0.0000) & 1.0000 (1.0000,0.0000) & \textbf{1.0000 (1.0000,0.0000)} &1.00 & 1.00 \\
  
  & GAN & 1.0000 (1.0000,0.0000) & 1.0000 (1.0000,0.0000) & \textbf{1.0000 (1.0000,0.0000)} &1.00 & 1.00 \\
  
 & SMOTified-GAN & 1.0000 (1.0000,0.0000) & 1.0000 (1.0000,0.0000) & \textbf{1.0000 (1.0000,0.0000)} &1.00 & 1.00 \\
 
 \hline
 Abalone  & Non-oversampled & 0.9080 (0.9108,0.0019)& 0.9072 (0.9114,0.0028) & 0.7556 (0.7658,0.0090) &0.80 & 0.73 \\ 
  
  & SMOTE & 0.8969 (0.9022,0.0035) & 0.8622 (0.8827,0.0105) & 0.7259 (0.7566,0.0200) & 0.78 & 0.72 \\
  
& GAN & 0.9422 (0.9439,0.0013) & 0.9070 (0.9125,0.0040) & 0.7555 (0.7687,0.0061) &0.80 & 0.74 \\
 
 & SMOTified-GAN & 0.9427 (0.9441,0.0008) & 0.9075 (0.9126,0.0036) & \textbf{0.7612 (0.7711,0.0065)} & 0.80 & \textbf{0.75} \\

  
  
 
  
  
 
  
  
 
 
    \hline  
 Spambase & Non-oversampled & 0.9476 (0.9527,0.0020)& 0.9309 (0.9380,0.0027) & 0.9152 (0.9213,0.0032) & 0.91 & 0.92 \\ 
  
  & SMOTE &0.9455 (0.9526,0.0026) & 0.9276 (0.9336,0.0031) & 0.9129 (0.9204,0.0049) & 0.92 & 0.92 \\

  & GAN & 0.9571 (0.9599,0.0019) & 0.9319 (0.9380,0.0030) & 0.9172 (0.9222,0.0036) &0.93 & 0.92 \\  
 
 & SMOTified-GAN & 0.9583 (0.9602,0.0012) & 0.9323 (0.9380,0.0026) & \textbf{0.9174 (0.9222,0.0031)} & \textbf{0.94} & 0.91 \\

   \hline
 Shuttle & Non-oversampled & 0.9994 (0.9998,0.0006)& 0.9992 (0.9996,0.0005) & 0.8256 (0.9350,0.2294) & 0.92 & 0.96 \\ 
  
  & SMOTE & 0.9996 (0.9996,0.0000) & 0.9990 (0.9993,0.0001) & 0.8465 (0.8837,0.0220) & 0.83 & \textbf{0.97} \\
 
  & GAN & 0.9995 (0.9999,0.0005) & 0.9989 (0.9996,0.0009) & 0.8497 (0.9367,0.2240) & 0.93 & 0.95 \\  
 
 & SMOTified-GAN & 0.9996 (0.9997,0.0004) & 0.9993 (0.9996,0.0006) & \textbf{0.8632 (0.9368,0.2009)} & 0.93 & 0.95 \\

 \hline
 Credit-Card & Non-oversampled & 0.9996  (0.9997,0.0001)& 0.9991  (0.9993,0.0001) & 0.8066  (0.8214,0.0327) & 0.84 & 0.80
 \\ 
  
 Fraud & SMOTE & 0.9996 (0.9997,0.0001) & 0.9990 (0.9991,0.0001) & 0.7099 (0.7409,0.0210) & 0.80 & 0.69 \\

 & GAN & 0.9995 (0.9997,0.0001) & 0.9991 (0.9994,0.0001) & 0.8069 (0.8214,0.0241) & 0.84 & 0.80 \\
 
 & SMOTified-GAN & 0.9993 (0.9994,0.0001) & 0.9992 (0.9993,0.0001) & \textbf{0.8118 (0.8243,0.0202)} & \textbf{0.85} & 0.80 \\
 
  \hline
 Connect4 & Non-oversampled & 0.9947 (0.9966,0.0014)& 0.9948 (0.9965,0.0013) & \textbf{0.9361 (0.9578,0.0151)} & 0.92 & 1.00 \\ 
  
  & SMOTE & 0.9967 (0.9970,0.0001) & 0.9912 (0.9930,0.0007) & 0.9011 (0.9167,0.0068) & 0.85 & 1.00 \\

  & GAN & 0.9962 (0.9982,0.0010) & 0.9938 (0.9965,0.0021) & 0.9251 (0.9577,0.0180)  & 0.92 & 1.00 \\ 
 
 & SMOTified-GAN & 0.9966 (0.9986,0.0009) & 0.9946 (0.9965,0.0017) & 0.9355 (0.9578,0.0158) & 0.92 & 1.00 \\

  \hline   
   \hline
\end{tabular}
\label{tab:results}
\end{table*}

\begin{figure*}[htbp!]\centering
\centering     
\subfigure[abalone-1]{\label{fig:AUC:abalone}\includegraphics[width=0.3\textwidth]{images/images of  curves/abalfi.png}}
\subfigure[creditcard]{\label{fig:AUC:creditcard}\includegraphics[width=0.3\textwidth]{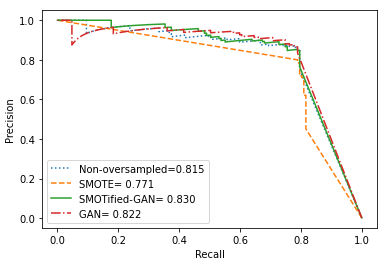}}
\subfigure[ecoli]{\label{fig:AUC:ecoli}\includegraphics[width=0.3\textwidth]{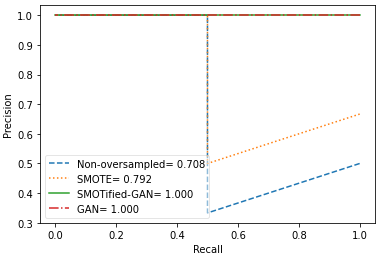}}
\subfigure[ionosphere]{\label{fig:AUC:ionosphere}\includegraphics[width=0.3\textwidth]{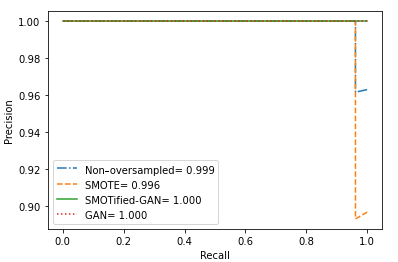}}
\subfigure[connect4]{\label{fig:AUC:connect4}\includegraphics[width=0.3\textwidth]{images/images of  curves/connfi.png}}
\subfigure[pageblocks]{\label{fig:AUC:pageblocks}\includegraphics[width=0.3\textwidth]{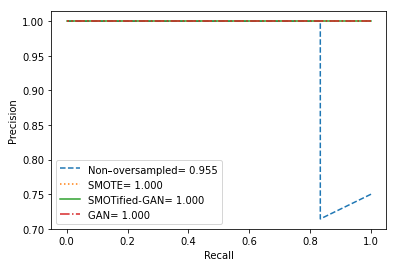}}
\subfigure[poker]{\label{fig:AUC:poker}\includegraphics[width=0.3\textwidth]{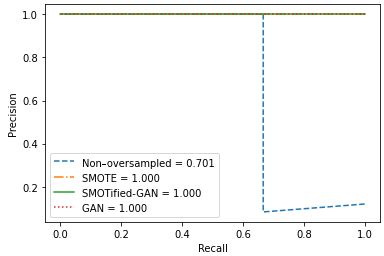}}
\subfigure[shuttle]{\label{fig:AUC:shuttle}\includegraphics[width=0.3\textwidth]{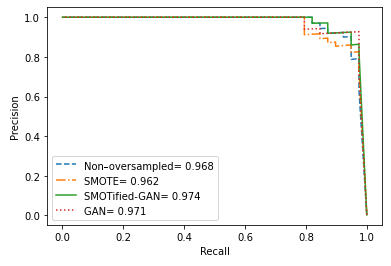}}
\subfigure[spambase]{\label{fig:AUC:spambase}\includegraphics[width=0.3\textwidth]{images/images of  curves/spamfi.png}}
\caption{Precision vs Recall for AUC Curves}
\label{fig:AUC}
\end{figure*}

\figurename{} \ref{fig:AUC} presents the receiver operating characteristic curve or ROC curve on precision and recall for the tested datasets. The results for nine datasets have been illustrated with all the tested algorithms. It also shows the measure for the area under the curve (AUC). This is a standard performance measure for imbalanced data. It is clear from the graph that our proposed SMOTified-GAN has highest AUC for all the datasets except Abalone. For example \figurename{} \ref{fig:AUC:shuttle} shows AUC for Shuttle dataset with SMOTified-GAN, GAN, no data augmentation and SMOTE have the result of 0.949, 0.911, 0.891 and 0.712 respectively in descending order. SMOTified-GAN is better than others by up to 0.038 to the next best algorithm. GAN and SMOTE shows the mixed results as discussed earlier with F1-scores.


 \section{Discussion}
A significant improvement in the quality of classification has been observed with the introduction of SMOTified-GAN as an oversampling technique. It has clearly outperformed na\"ive GAN and SMOTE in most of the datasets. The F1-score has been improved by up to 9\% for Ecoli dataset from the next best oversampling technique where the precision has also shown significant growth of around \{7\% to 8\%\} for Wine and Yeast datasets. The recall has not been much improved. SMOTE generally has low precision and SMOTified-GAN has relatively better precision than the other models. Most notable improvement from GAN and SMOTE can be seen with \{Abalone, Pageblock, Wine, and Shuttle datasets\}, and \{Credit-card Fraud and Wine datasets\} respectively.

Considering the impact of the number of features, percentage of minority class and the size of the data on the quality of oversampling has shown mixed results. For example datasets with higher features such as Credit-card fraud (30 features, 0.172\% of minority class) have shown good results but Connect4 (42 features, 3.84\% of minority class) has shown poorer results. However, SMOTE generally performs poorly on F1-score even with raw data in 3/4 times when the features are high as in Connect4 (42 features), Credit-card fraud (30 features), Ionosphere (34 features) and Spambase (57 features).

Furthermore, the best algorithm may not be clearly visible with ROC curves in \figurename{} \ref{fig:AUC}, however, the AUC-ROC measures for each graph shows that SMOTified-GAN outperforms other algorithms. The larger area the ROC curve occupies the better the algorithm which is shown by the AUC measures. For example, it is somewhat clear from the Shuttle that shows the best to worst in the order of SMOTified-GAN  (0.949), GAN  (0.911), no oversampling technique  (0.891) and then SMOTE  (0.712). So SMOTified-GAN is 3.5\% better than the next best algorithm. Similarly, it is 2.1\% better than the second best algorithm for Spambase.


 \section{Conclusion}
 
 We presented a framework that addressed class imbalanced pattern classification problems by combining features from GAN and SMOTE. Our results show that the proposed framework significantly improves the majority of the class imbalanced problems. There were improvements of up to 9\% on the F1 score for the benchmark datasets. Since it is an offline pre-processing technique with a reasonable time complexity order of $O(N^2d^2T)$, it does not affect the efficiency of the training process. We also visualized the learning process and found out that the AUC of SMOTified-GAN is better than the $2^{nd}$ best algorithm up to  2.1\% (for Spambase) and 3.5\% (for Shuttle). 
 
 There are several possible future directions from this work such as applying SMOTified-GAN to other neural networks such as CNNs and recurrent neural networks (RNNs) to oversample imbalanced image datasets and time-series data, respectively.  Furthermore, different variations and combinations of SMOTE and GAN for the new model of SMOTified-GAN can improve it even further.
 
 It will be interesting to investigate the conjoining of GAN with other over-sampling techniques such as MCMC. Its sampling method on a Bayesian framework can be used to incorporate uncertainty in the predictions and develop a probabilistic data generation process via GANs. The proposed framework can be used in a wide range of problems that face challenges when it comes to class imbalance issues. This framework can also be used to improve few-shot learning \cite{wang_generalizing_2020} to address problems where the model finds it difficult to draw decision boundaries due to a lack of data. Moreover, we can also investigate if the method can be used to address the bias-variance problems in order to improve the generalization ability of the model given that the training data differs significantly from the test dataset. 
 

 \section*{Code and Data}

We provide Python code and data   for extending this work further \footnote{ \url{https://github.com/sydney-machine-learning/GANclassimbalanced}}.








\bibliographystyle{IEEEtran}
\bibliography{access.bib}

\EOD

\end{document}